\documentclass[runningheads,a4paper]{llncs}

\usepackage{amssymb}
\usepackage{graphicx}

\usepackage{url}
\urldef{\mailsa}\path|{alfred.hofmann, ursula.barth, ingrid.haas, frank.holzwarth,|
\urldef{\mailsb}\path|anna.kramer, leonie.kunz, christine.reiss, nicole.sator,|
\urldef{\mailsc}\path|erika.siebert-cole, peter.strasser, lncs}@springer.com|
\newcommand{\keywords}[1]{\par\addvspace\baselineskip
\noindent\keywordname\enspace\ignorespaces#1}

\usepackage{subfigure}
\usepackage{multirow}
\usepackage{color}

\begin{document}

\mainmatter  

\title{Adaptive Measurement Network for CS \\Image Reconstruction\footnotetext{This paper will be presented at the CCF Chinese Conference on Computer Vision, Tianjin, 11-14, October, 2017.}
}

\titlerunning{Adaptive Measurement for CS}

%
%
\author{Xuemei Xie%
\and Yuxiang Wang\and Guangming Shi\and Chenye Wang\and\\
Jiang Du\and Zhifu Zhao}
\authorrunning{Xuemei Xie et al.}

\institute{Xidian University, Xi'an, China\\
\email{xmxie@mail.xidian.edu.cn}}

%
%

\toctitle{Adaptive Measurement Network for CS Image Reconstruction}
\tocauthor{Xuemei Xie et al.}
\maketitle

\begin{abstract}
Conventional compressive sensing (CS) reconstruction is very slow for its characteristic of solving an optimization problem. Convolutional neural network can realize fast processing while achieving comparable results. While CS image recovery with high quality not only depends on good reconstruction algorithms, but also good measurements. In this paper, we propose an adaptive measurement network in which measurement is obtained by learning. The new network consists of a fully-connected layer and ReconNet. The fully-connected layer which has low-dimension output acts as measurement. We train the fully-connected layer and ReconNet simultaneously and obtain adaptive measurement. Because the adaptive measurement fits dataset better, in contrast with random Gaussian measurement matrix, under the same measurement rate, it can extract the information of scene more efficiently and get better reconstruction results. Experiments show that the new network outperforms the original one.
\end{abstract}

\section{Introduction}

Compressive sensing (CS) theory\cite{n1,n2,n3} is able to acquire measurements of signal at sub-Nyquist rates and recover signal with high probability when the signal is sparse in a certain domain. Random Gaussian matrix is often used as the measurement matrix because we must ensure that the basis of sparse domain is incoherent with measurement. When it comes to reconstruction, there are two main kinds of reconstruction methods: conventional reconstruction methods\cite{n4,n5,n6,n7,n8,n9} and deep learning reconstruction methods \cite{n10,n11,n12,n13}.

A large amount of compressive sensing reconstruction methods have been proposed. But almost all of them get the reconstruction result by solving optimization, which makes them slow. In recent years, some deep learning approaches have been proposed. With its characteristic of off-line training and online test, the speed of reconstruction has been greatly improved.

The first paper\cite{n11} applying deep learning approach to solve the CS recovery problem used stacked denoising autoencoders (SDA) to recover signals from undersampled measurements. SDA consists of fully-connected layers, which means larger network with the signal size growing. This imposes a large computational complexity and can lead to overfitting. DeepInverse\cite{n13}, utilizing convolutional neuron network (CNN) layers, works with arbitrary measurement, which means that the whole image can be reconstructed by it. But with the signal size growing, the cost of measurement grows too. ReconNet\cite{n10} used a fully-connected layer along with convolutional layers to recover signals from compressive measurements block-wise. It can reduce the network complexity and the training time while ensuring a good reconstruction quality.However, ReconNet used fixed random Gaussian measurement, which is not optimally designed for signal.

In this paper, we take a fully-connected layer which has low-dimension output as measurement. The fully-connected layer and reconstruction network ReconNet are put together to be an adaptive measurement network. It can be proved the adaptive measurement network performs better than ReconNet with fixed measurement by comparing the trained weights. Experiment shows that the adaptive measurement matrix can obtain more information of images than fixed random Gaussian measurement matrix. And the images reconstructed from adaptive measurement have larger value of PSNR.

The structure of the paper is organized as follows. Section\,2 introduces the fixed random Gaussian measurement. And the description of adaptive measurement network is introduced in Section\,3. Section\,4 conducts the experiments, and Section\,5 concludes the paper.
\section{Fixed random Gaussian measurement}
\begin{figure}
\centering
\includegraphics[height=5cm, width=12cm]{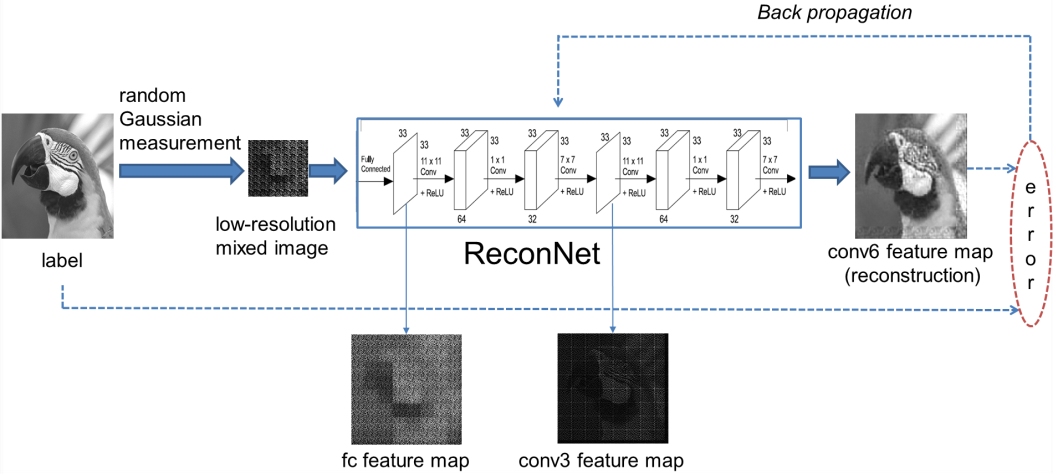}
\caption{The process of training and testing ReconNet with fixed random Gaussian measurement. Feature maps of fully connected layer (fc feature map), third convolutional layer (conv3 feature map) and sixth convolutional layer (conv6 feature map) at measurement rate 25\%.}
\label{fig:fig1}
\end{figure}

ReconNet is a deep learning CS reconstruction approach, which can recover rich semantic content even at a low measurement rate of 1\%. It is a convolutional neural network and the process of training and testing ReconNet are shown in Fig.\,\ref{fig:fig1}.

ReconNet consists of one fully-connected layer and six convolutional layers in which the first three layers and last three layers are identical. The function of these layers is described as follows. The fully-connected layer takes CS measurements as input and outputs a feature map of size 33x33. The first/last three layers are inspired by SRCNN\cite{n14}, which is a CNN-based approach for image super-resolution. Except for the last convolutional layer, all the other convolutional layers followed by ReLU. Only when the input of CNN has structure information can CNN work. So the fully-connected layer plays a role of recovering some structure information from CS measurements and then convolutional layers enhance output of full-connected layer to a high-resolution image.

The training dataset consists of input data and ground truth. All the images in dataset are 33x33 size patches extracted from original images. Input data of ReconNet for training is obtained by measuring each of the extracted patches using a random Gaussian matrix $\Phi$. For a given measurement rate, a random Gaussian matrix of appropriate size is firstly generated and then its rows are orthonormalized to get $\Phi$. The input of testing and training is obtained by using the same random Gaussian matrix. Before being measured, the 33x33 size block should be reshaped into a 1089-dimension column vector.

The loss function is given by
$$L(\{ W\} ) = \;\frac{1}{T}\sum\nolimits_i^T {{{\left\| {f({y_i},\{ W\} ) - \;{x_i}} \right\|}^2}}  . \eqno{(1)}$$
$f({y_i},\{ W\})$ is the $i$--th reconstruction image of ReconNet, ${x_i}$ is the $i$--th original signal as well as the $i$--th label, $W$ means all parameters in ReconNet. $T$ is the total number of image blocks in the training dataset. The loss function is minimized by adjusting $W$ using backpropagation. For each measurement rate, two networks are trained, one with random Gaussian initialization for the fully connected layer, and the other with a deterministic initialization, in each case, weights of all convolutional layers are initialized using a random Gaussian with a fixed standard deviation. The network which provides the lower loss on a validation test will be chosen.

The test process does not include the dotted line part in Fig.\,\ref{fig:fig1}. The high-resolution scene image is divided into non-overlapping blocks of size 33x33 and each of them is reconstructed by feeding in the corresponding CS measurements to ReconNet. The reconstructed blocks are arranged appropriately to form a reconstruction of the image.

\begin{figure}
\centering
\includegraphics[height=5cm, width=12cm]{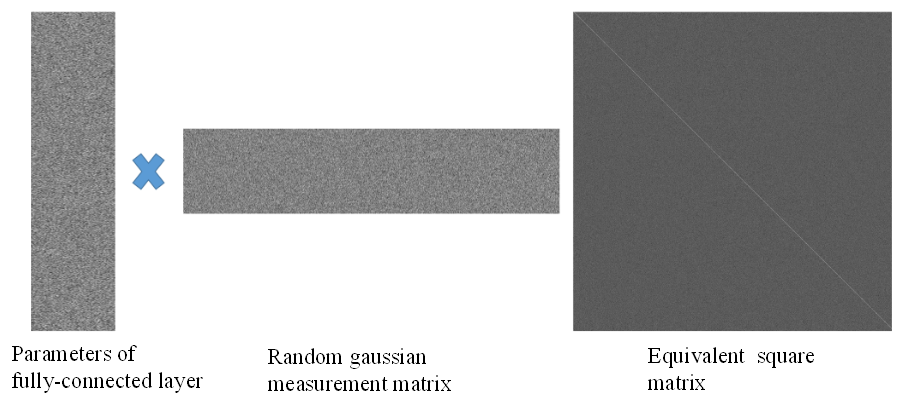}
\caption{The equivalent square matrix at measurement rate 10\%.}
\label{fig:fig2}
\end{figure}

It can be proved that fully-connected layer can recover some structure information. The block of high-resolution scene image is firstly measured by random Gaussian matrix and then multiply parameters of fully-connected layer, this process equals to multiplying a square matrix as Fig.\,\ref{fig:fig2} shows. The diagonal numbers of square matrix are obviously larger than the other numbers, and we can see it as an approximate unit matrix, which means the $i$--th element of output of fully-connected layer is mainly determined by the $i$--th element of high resolution block. Feature maps can also be used to prove that the fully-connected layer can recover some information.

The feature map of fully-connected layer (fc feature map in Fig.\,\ref{fig:fig1}) can be obtained after image parrot (label in Fig.\,\ref{fig:fig1}) is sent into the trained ReconNet at measurement rate 10\%. It can be seen that the fully-connected layer can recover some structure information. The feature maps of fully-connected layer (fc feature map in Fig.\,\ref{fig:fig1}), third convolutional layer (conv3 feature map in Fig.\,\ref{fig:fig1}) and sixth convolutional layer (conv6 feature map in Fig.\,\ref{fig:fig1}) at measurement rate 25\% are also shown in Fig.\,\ref{fig:fig1}. We can see that it is a process from low-resolution to high-resolution.

The main drawback of random measurement is that they are not optimally designed
for signal. Therefore, the adaptive measurement is possibly a promising approach.

\section{Adaptive measurement network}
\begin{figure}
\centering
\includegraphics[height=5cm, width=12cm]{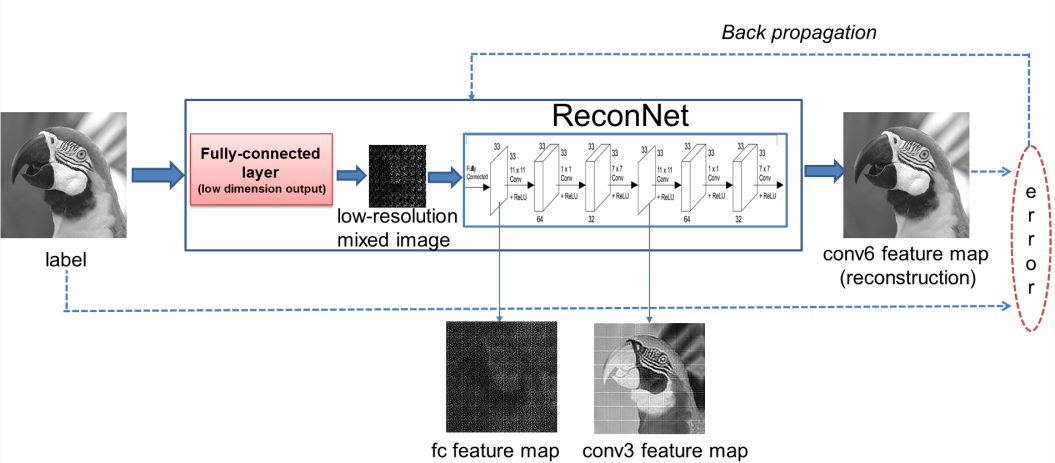}
\caption{The process of training and testing with adaptive measurement. Feature maps of fully connected layer(fc feature map), third convolutional layer(conv3 feature map) and sixth convo-lutional layer(conv6 feature map) at measurement rate 10\%.}
\label{fig:fig3}
\end{figure}

We put a fully-connected layer and ReconNet together to form the adaptive measurement
network as Fig.\,\ref{fig:fig3} shows. The fully-connected layer which has low-dimension
output is considered as measurement. The measurement rate is determined by dimension
of network input and fully-connected layer output.

The whole Fig.\,\ref{fig:fig3} shows the process of training adaptive network (including the
dashed line part). In the training stage, ground truth still consists of 33x33 size patches
extracted from the original images. Different from the random Gaussian measurement
network, input data of training set is the same with ground truth instead of the
output of Gaussian measurements.

When it comes to test, the parameter of fully-connected layer is taken as measurement
while ReconNet still works as reconstruction network. Fig.\,\ref{fig:fig3} without dashed
line part shows the process of testing.

Accordingly, the loss function of new network is given by
$$L(\{ W\} ) = \;\frac{1}{T}\sum\nolimits_i^T {{{\left\| {f({x_i},\{ W,K\} ) - \;{x_i}} \right\|}^2}}  . \eqno{(2)}$$
where $K$ is the parameter of new added fully-connected layer. Difference between (1)
and (2) is that in (2) reconstruction image is determined by ${x_i}$ and $\{ W,K\}$, but ${y_i}$ and $\{ W\}$ in (1).

Compared to the original one, the new network has more parameter to train. The initial value of the network is random Gaussian. There is a high probability that a better measurement more adaptive to data set can be obtained.

\begin{figure}
\centering

\subfigure[]{
\label{fig:fig4:a} 
\includegraphics[height=3.9cm, width=3.9cm]{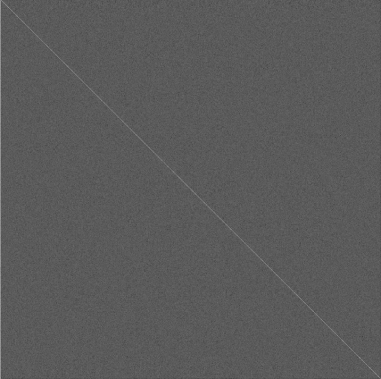}}
\hspace{1in}
\subfigure[]{
\label{fig:fig4:b} 
\includegraphics[height=3.9cm, width=3.9cm]{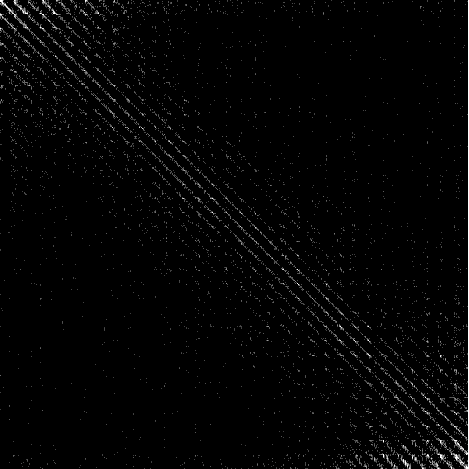}}
\caption{The equivalent square matrix (a) Fixed random Gaussian measurement with MR=10\% and (b) adaptive measurement with MR=10\%.}
\label{fig:fig4} 

\end{figure}

\begin{figure}
\centering
\includegraphics[height=4.9cm, width=11.5cm]{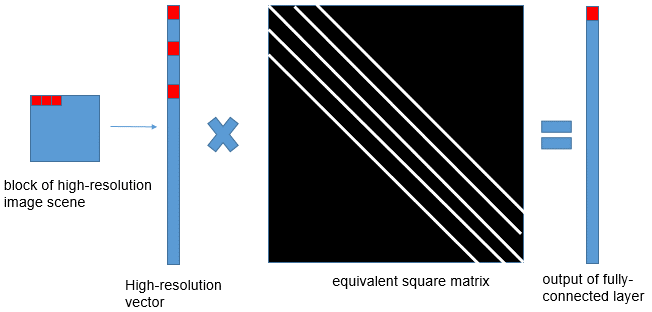}
\caption{Simplified explanation.}
\label{fig:fig5}
\end{figure}

It can be proved the new ReconNet of adaptive measurement network performs better than the original one. The fully-connected layer can recover more structure information. The equivalent process is shown in Fig.\,\ref{fig:fig4:b}. Compared with Fig.\,\ref{fig:fig4:a}, the value of square matrix in Fig.\,\ref{fig:fig4:b} is more dispersing. The $i$--th element of output of fully-connected layer is mainly determined by the $i$--th element of high resolution block and its neighboring elements. So, the measurement can acquire information more effectively. As shown in Fig.\,\ref{fig:fig5}, in equivalent square matrix, the value of white part is larger than black part. A high-resolution block is firstly reshaped to a high-resolution vector. The output of fully-connected layer is obtained by multiplying the high-resolution vector by the equivalent square matrix. We take the 1--th element as an example. The 1--th element of output is obtained by multiplying the high-resolution vector by 1--th row vector of square matrix. The elements of 1--th row vector can be seen as the weights of column vector. It is obvious that the red elements of the high-resolution vector have larger weights, which means the 1--th element of output of fully-connected layer is mainly determined by those red elements of high-resolution vector. In high-resolution block, those red elements correspond to the 1--th element and its neighboring elements. Since the 1--th element and its neighboring elements are relevant, the values of them are approximate, which means the 1-th element of output of fully-connected layer is more determined by the 1--th element of high-resolution block. The $i$--th element can be explained the same way. It is proved that the fully-connected layer of new ReconNet recover more structure information.

Adaptive measurement network's feature maps of fully-connected layer (fc feature map), third convolutional layer (conv3 feature map) and sixth convolutional layer (conv6 feature map) at measurement rate 10\% are shown in Fig.\,\ref{fig:fig3}. In contrast to Fig.\,\ref{fig:fig1}, the feature maps of adaptive network are obviously better even at measurement 10\%.

\begin{figure}
\centering

\subfigure[]{
\label{fig:fig6:a} 
\includegraphics[height=3.5cm]{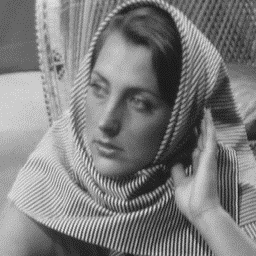}}
\subfigure[]{
\label{fig:fig6:b} 
\includegraphics[height=3.5cm]{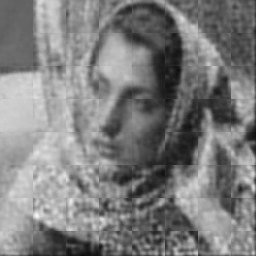}}
\subfigure[]{
\label{fig:fig6:c} 
\includegraphics[height=3.5cm]{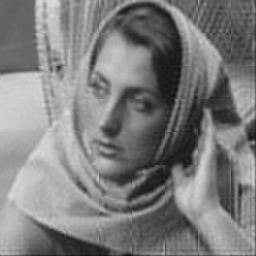}}
\caption{Test image and reconstruction results of Barbara. (a) the original image (b) the reconstruction result of random Gaussian measurement network (c) the reconstruction result of adaptive measurement network.}
\label{fig:fig6} 

\end{figure}

Fig.\,\ref{fig:fig6} shows an example of reconstruction results at two kinds of measurement. The measurement rate is 10\%. Fig.\,\ref{fig:fig6:a} is the original image. Fig.\,\ref{fig:fig6:b} is the reconstruction result of random Gaussian measurement network. Fig.\,\ref{fig:fig6:c} is the reconstruction result of adaptive network. The adaptive reconstruction result is more attractive visually.

\begin{figure}
\centering
\subfigure[]{
\label{fig:fig7:a}
\begin{minipage}[a]{0.5\textwidth}
\centerline{\includegraphics[height=3cm, width=7cm]{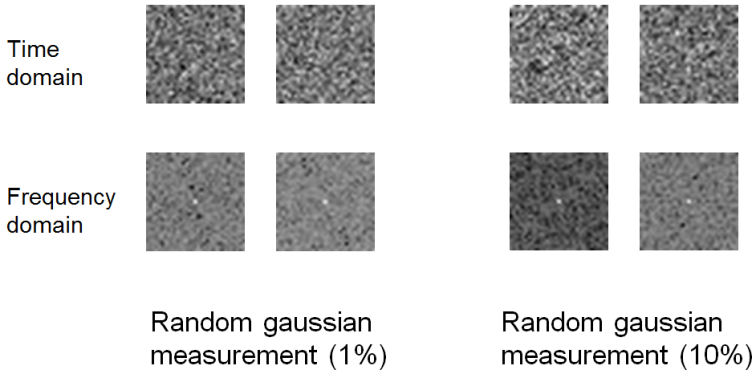}}
\end{minipage}
}%

\subfigure[]{
\label{fig:fig7:b}
\begin{minipage}[b]{0.8\textwidth}
\centerline{\includegraphics[height=3cm, width=10cm]{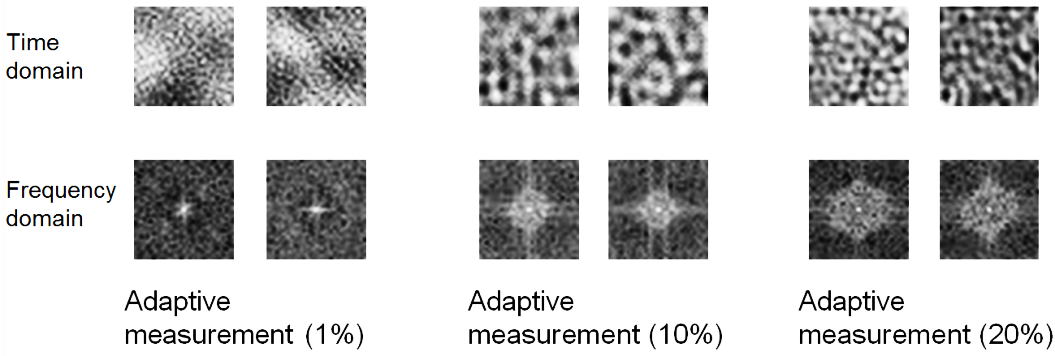}}
\end{minipage}
}

\caption{Measurement matrix. (a) is random Gaussian measurement matrix at measurement rate 1\%,10\% in time and frequency domain. (b) is adaptive measurement matrix at measurement rate 1\%, 10\%, 20\% in time and frequency domain.}
\label{fig:fig7} 

\end{figure}

Adaptive measurement network's better performance can also be proved through measurement matrix. Since the original signal is reshaped to a column vector before being measured, we reshape some row vectors of measurement matrix to size 33x33. Two reshaped row vectors of the random Gaussian measurement matrix at measurement rate 1\% and 10\% in time and frequency domain are shown in Fig.\,\ref{fig:fig7:a}. The content of random Gaussian measurement matrix is obviously irregular. We cannot get any useful information from Fig.\,\ref{fig:fig7:a}. Two reshaped row vectors of adaptive measurement matrix at measurement rate 1\%, 10\%, 20\% in time and frequency domain are shown in Fig.\,\ref{fig:fig7:b}. As we all know, most of the energy of an image is concentrated in the low frequency part. When the measurement rate is low, some high frequency information must be discarded to reconstruct the contours of the image as fully as possible. However, with the increase in measurement rate, the ability of measurement is enhanced. The high-frequency information in adaptive measurement increases gradually. We can also know it from frequency domain image. So, the reconstructed image will become clearer.

\section{Results}
In this section, we conduct reconstruction experiments at both fixed random Gaussian measurement and adaptive measurement.

We use the caffe framework for network training on the MATLAB platform. Our computer is equipped with Intel Core i7-6700 CPU with frequency of 3.4GHz, NVidia GeForce GTX 980 GPU, 64GB RAM, and the framework runs on the Ubuntu 14.04 operating system.

The dataset consists of 21760 33x33 size patches extracted from 91 images in\cite{n14} with a stride equal to 14. It is worthy to mention that because ReconNet reconstruct image block-wise and the size of block is fixed, zero-padding operation is applied to input images of different size. But we find symmetric padding acts better than zero-padding, so all the experiment results are based on symmetric padding instead of zero-padding.

We use cameraman image to test both the networks, and the result of reconstruction is shown as follows.

\begin{figure}
\centering
\includegraphics[height=6cm, width=10cm]{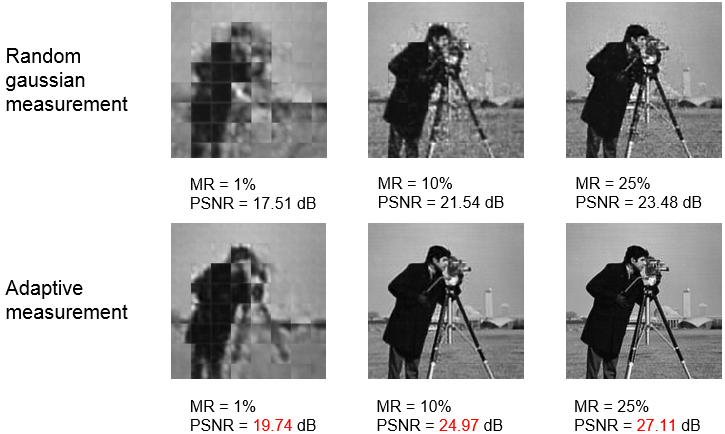}
\caption{The reconstruction results of image cameraman at different measurement rates.}
\label{fig:fig8}
\end{figure}

It is shown in Fig.\,\ref{fig:fig8} that the reconstruction results of image cameraman at different measurement rates with different measurements. Our results are more attractive visually.

The reconstruction results for 11 test images at measurement rate 1\%, 10\%, 25\% with different measurements are shown in Table\,\ref{tab:tab1}. All results show that adaptive measurement outperforms random Gaussian measurement.

\begin{table}
\centering
\caption{The reconstruction results for 11 test images at measurement rate 1\%, 10\%, 25\% with different measurements.}
\begin{tabular}{cccccccccc}
\hline\noalign{\smallskip}
Image&
\multicolumn{2}{c}{Measurement\quad\quad} &
\multicolumn{2}{c}{Rate 25\%\quad\quad} &
\multicolumn{2}{c}{Rate 10\%\quad\quad} &
\multicolumn{2}{c}{Rate 1\%\quad\quad} \\
\noalign{\smallskip}
\hline\noalign{\smallskip}
\multirow{2}{*}{Monarch} &
\multicolumn{2}{c}{Adaptive\quad\quad} &
\multicolumn{2}{c}{{\color{red} 29.25dB}\quad\quad} &
\multicolumn{2}{c}{{\color{red} 26.65dB}\quad\quad} &
\multicolumn{2}{c}{{\color{red} 17.70dB}\quad\quad}  \\ &
\multicolumn{2}{c}{Gaussian\quad\quad} &
\multicolumn{2}{c}{24.95dB\quad\quad} &
\multicolumn{2}{c}{21.49dB\quad\quad} &
\multicolumn{2}{c}{15.61dB\quad\quad} \\
\noalign{\smallskip}
\hline
\noalign{\smallskip}
\multirow{2}{*}{Parrots} &
\multicolumn{2}{c}{Adaptive\quad\quad} &
\multicolumn{2}{c}{{\color{red} 30.51dB}\quad\quad} &
\multicolumn{2}{c}{{\color{red} 27.59dB}\quad\quad} &
\multicolumn{2}{c}{{\color{red} 21.67dB}\quad\quad}  \\ &
\multicolumn{2}{c}{Gaussian\quad\quad} &
\multicolumn{2}{c}{26.66dB\quad\quad} &
\multicolumn{2}{c}{23.36dB\quad\quad} &
\multicolumn{2}{c}{18.93dB\quad\quad} \\
\noalign{\smallskip}
\hline
\noalign{\smallskip}
\multirow{2}{*}{Barbara} &
\multicolumn{2}{c}{Adaptive\quad\quad} &
\multicolumn{2}{c}{{\color{red} 27.40dB}\quad\quad} &
\multicolumn{2}{c}{{\color{red} 24.28dB}\quad\quad} &
\multicolumn{2}{c}{{\color{red} 21.36dB}\quad\quad}  \\ &
\multicolumn{2}{c}{Gaussian\quad\quad} &
\multicolumn{2}{c}{23.58dB\quad\quad} &
\multicolumn{2}{c}{22.17dB\quad\quad} &
\multicolumn{2}{c}{19.08dB\quad\quad} \\
\noalign{\smallskip}
\hline
\noalign{\smallskip}
\multirow{2}{*}{Boats} &
\multicolumn{2}{c}{Adaptive\quad\quad} &
\multicolumn{2}{c}{{\color{red} 32.47dB}\quad\quad} &
\multicolumn{2}{c}{{\color{red} 28.80dB}\quad\quad} &
\multicolumn{2}{c}{{\color{red} 21.09dB}\quad\quad}  \\ &
\multicolumn{2}{c}{Gaussian\quad\quad} &
\multicolumn{2}{c}{27.83dB\quad\quad} &
\multicolumn{2}{c}{24.56dB\quad\quad} &
\multicolumn{2}{c}{18.82dB\quad\quad} \\
\noalign{\smallskip}
\hline
\noalign{\smallskip}
\multirow{2}{*}{Cameraman} &
\multicolumn{2}{c}{Adaptive\quad\quad} &
\multicolumn{2}{c}{{\color{red} 27.11dB}\quad\quad} &
\multicolumn{2}{c}{{\color{red} 24.97dB}\quad\quad} &
\multicolumn{2}{c}{{\color{red} 19.74dB}\quad\quad}  \\ &
\multicolumn{2}{c}{Gaussian\quad\quad} &
\multicolumn{2}{c}{23.48dB\quad\quad} &
\multicolumn{2}{c}{21.54dB\quad\quad} &
\multicolumn{2}{c}{17.51dB\quad\quad} \\
\noalign{\smallskip}
\hline
\noalign{\smallskip}
\multirow{2}{*}{Fingerprint} &
\multicolumn{2}{c}{Adaptive\quad\quad} &
\multicolumn{2}{c}{{\color{red} 32.31dB}\quad\quad} &
\multicolumn{2}{c}{{\color{red} 26.55dB}\quad\quad} &
\multicolumn{2}{c}{{\color{red} 16.22dB}\quad\quad}  \\ &
\multicolumn{2}{c}{Gaussian\quad\quad} &
\multicolumn{2}{c}{26.15dB\quad\quad} &
\multicolumn{2}{c}{20.99dB\quad\quad} &
\multicolumn{2}{c}{15.01dB\quad\quad} \\
\noalign{\smallskip}
\hline
\noalign{\smallskip}
\multirow{2}{*}{Flinstones} &
\multicolumn{2}{c}{Adaptive\quad\quad} &
\multicolumn{2}{c}{{\color{red} 27.94dB}\quad\quad} &
\multicolumn{2}{c}{{\color{red} 23.83dB}\quad\quad} &
\multicolumn{2}{c}{{\color{red} 16.12dB}\quad\quad}  \\ &
\multicolumn{2}{c}{Gaussian\quad\quad} &
\multicolumn{2}{c}{22.74dB\quad\quad} &
\multicolumn{2}{c}{19.04dB\quad\quad} &
\multicolumn{2}{c}{14.14dB\quad\quad} \\
\noalign{\smallskip}
\hline
\noalign{\smallskip}
\multirow{2}{*}{Foreman} &
\multicolumn{2}{c}{Adaptive\quad\quad} &
\multicolumn{2}{c}{{\color{red} 36.18dB}\quad\quad} &
\multicolumn{2}{c}{{\color{red} 33.51dB}\quad\quad} &
\multicolumn{2}{c}{{\color{red} 25.53dB}\quad\quad}  \\ &
\multicolumn{2}{c}{Gaussian\quad\quad} &
\multicolumn{2}{c}{32.08dB\quad\quad} &
\multicolumn{2}{c}{29.02dB\quad\quad} &
\multicolumn{2}{c}{22.03dB\quad\quad} \\
\noalign{\smallskip}
\hline
\noalign{\smallskip}
\multirow{2}{*}{House} &
\multicolumn{2}{c}{Adaptive\quad\quad} &
\multicolumn{2}{c}{{\color{red} 34.38dB}\quad\quad} &
\multicolumn{2}{c}{{\color{red} 31.43dB}\quad\quad} &
\multicolumn{2}{c}{{\color{red} 22.93dB}\quad\quad}  \\ &
\multicolumn{2}{c}{Gaussian\quad\quad} &
\multicolumn{2}{c}{29.96dB\quad\quad} &
\multicolumn{2}{c}{26.74dB\quad\quad} &
\multicolumn{2}{c}{20.30dB\quad\quad} \\
\noalign{\smallskip}
\hline
\noalign{\smallskip}
\multirow{2}{*}{Lena} &
\multicolumn{2}{c}{Adaptive\quad\quad} &
\multicolumn{2}{c}{{\color{red} 31.63dB}\quad\quad} &
\multicolumn{2}{c}{{\color{red} 28.50dB}\quad\quad} &
\multicolumn{2}{c}{{\color{red} 21.49dB}\quad\quad}  \\ &
\multicolumn{2}{c}{Gaussian\quad\quad} &
\multicolumn{2}{c}{27.47dB\quad\quad} &
\multicolumn{2}{c}{24.48dB\quad\quad} &
\multicolumn{2}{c}{18.51dB\quad\quad} \\
\noalign{\smallskip}
\hline
\noalign{\smallskip}
\multirow{2}{*}{Peppers} &
\multicolumn{2}{c}{Adaptive\quad\quad} &
\multicolumn{2}{c}{{\color{red} 29.65dB}\quad\quad} &
\multicolumn{2}{c}{{\color{red} 26.67dB}\quad\quad} &
\multicolumn{2}{c}{{\color{red} 19.75dB}\quad\quad}  \\ &
\multicolumn{2}{c}{Gaussian\quad\quad} &
\multicolumn{2}{c}{25.74dB\quad\quad} &
\multicolumn{2}{c}{22.72dB\quad\quad} &
\multicolumn{2}{c}{17.39dB\quad\quad} \\
\noalign{\smallskip}
\hline
\noalign{\smallskip}
\multirow{2}{*}{MeanPSNR} &
\multicolumn{2}{c}{Adaptive\quad\quad} &
\multicolumn{2}{c}{{\color{red} 30.80dB}\quad\quad} &
\multicolumn{2}{c}{{\color{red} 27.53dB}\quad\quad} &
\multicolumn{2}{c}{{\color{red} 20.33dB}\quad\quad}  \\ &
\multicolumn{2}{c}{Gaussian\quad\quad} &
\multicolumn{2}{c}{26.42dB\quad\quad} &
\multicolumn{2}{c}{23.28dB\quad\quad} &
\multicolumn{2}{c}{17.94dB\quad\quad} \\
\noalign{\smallskip}
\hline
\end{tabular}
\label{tab:tab1}
\end{table}

\section{Conclusion}
We have presented an adaptive measurement obtained by learning. We showed that the adaptive measurement provides better reconstruction results than the fixed random Gaussian measurement. It is shown that the learned measurement matrix is more regular in time domain. It is clear that the learned measurement matrix is more adaptive to data set than the fixed one. That's an important reason why adaptive measurement works better. What's more, our network is universal, which can be applied to all kinds of images.

\section{Acknowledgements}
This work is supported by the National Natural Science Foundation of China (Grant No.61472301, 61632019) and the Foundation for Innovative Research Groups of the National Natural Science Foundation of China(No.61621005).

\end{document}